\documentclass[letterpaper]{article} 
\usepackage{aaai24}  
\usepackage{times}  
\usepackage{helvet}  
\usepackage{courier}  
\usepackage[hyphens]{url}  
\usepackage{graphicx} 
\usepackage{natbib}  
\usepackage{caption} 
\usepackage{algorithm}

\usepackage{newfloat}
\usepackage{listings}
\DeclareCaptionStyle{ruled}{labelfont=normalfont,labelsep=colon,strut=off} 
\floatstyle{ruled}
\newfloat{listing}{tb}{lst}{}
\floatname{listing}{Listing}
\nocopyright 

\title{Memory-adaptive Depth-wise Heterogeneous Federated Learning}

\author{
    Kai Zhang \textsuperscript{\rm 1*}, Yutong Dai \textsuperscript{\rm 1}, Hongyi Wang \textsuperscript{\rm 2}, Eric Xing \textsuperscript{\rm 2,4,5}, Xun Chen \textsuperscript{\rm 3}, Lichao Sun \textsuperscript{\rm 1}
}
\affiliations{
    \textsuperscript{\rm 1 }Lehigh University, \textsuperscript{\rm 2 }Carnegie Mellon University, 
    \textsuperscript{\rm 3 }Samsung Research America, \\
    \textsuperscript{\rm 4 }Mohamed bin Zayed University of Artificial Intelligence,  
    \textsuperscript{\rm 5 }Petuum Inc. \\
    \textsuperscript{*}kaz321@lehigh.edu
}

\usepackage{bibentry}

\usepackage{amsmath}
\usepackage{amssymb}
\usepackage{booktabs}
\usepackage{caption}
\usepackage{multirow}
\usepackage{xcolor}
\usepackage{soul}
\usepackage{booktabs, arydshln}  
\usepackage{algpseudocode}  
\makeatletter
\def\adl@drawiv#1#2#3{%
        \hskip.5\tabcolsep
        \xleaders#3{#2.5\@tempdimb #1{1}#2.5\@tempdimb}%
                #2\z@ plus1fil minus1fil\relax
        \hskip.5\tabcolsep}
\newcommand{\cdashlinelr}[1]{%
  \noalign{\vskip\aboverulesep
           \global\let\@dashdrawstore\adl@draw
           \global\let\adl@draw\adl@drawiv}
  \cdashline{#1}
  \noalign{\global\let\adl@draw\@dashdrawstore
           \vskip\belowrulesep}}
\makeatother

\usepackage{notations}
\DeclareMathOperator{\KL}{\boldsymbol{\mathbf{KL}}}

\usepackage{colortbl}
\definecolor{Gray}{gray}{0.9}
\definecolor{LightCyan}{rgb}{0.88,1,1}
\definecolor{OliveGreen}{rgb}{0.427,0.443,0.180}

\newcommand{\fedepth}{\textsc{FeDepth}}

\begin{document}

\maketitle

\begin{abstract}
Federated learning is a promising paradigm that allows multiple clients to collaboratively train a model without sharing the local data. However, the presence of heterogeneous devices in federated learning, such as mobile phones and IoT devices with varying memory capabilities, would limit the scale and hence the performance of the model could be trained. The mainstream approaches to address memory limitations focus on width-slimming techniques, where different clients train subnetworks with reduced widths locally and then the server aggregates the subnetworks. The global model produced from these methods suffers from performance degradation due to the negative impact of the actions taken to handle the varying subnetwork widths in the aggregation phase. 
In this paper, we introduce a memory-adaptive depth-wise learning solution in FL called \fedepth{}, which adaptively decomposes the full model into blocks according to the memory budgets of each client and trains blocks sequentially to obtain a full inference model. 
Our method outperforms state-of-the-art approaches, achieving 5\% and more than 10\% improvements in top-1 accuracy on CIFAR-10 and CIFAR-100, respectively.
We also demonstrate the effectiveness of depth-wise fine-tuning on ViT.
Our findings highlight the importance of memory-aware techniques for federated learning with heterogeneous devices and the success of depth-wise training strategy in improving the global model's performance. 
\end{abstract}

\section{Introduction}

Federated Learning (FL) is a popular distributed learning paradigm that can address decentralized data and privacy-preserving challenges by collaboratively training a model among multiple local clients without centralizing their private data \cite{mcmahan2017communication, kairouz2021advances}. FL has gained widespread interest and has been applied in numerous applications, such as healthcare~\cite{du2022flamby}, anomaly detection~\cite{zhang2021federated}, recommendation system~\cite{lin2020meta}, and knowledge graph completion~\cite{zhang-etal-2022-efficient-federated}. However, a defining trait of FL is the presence of heterogeneity \textemdash{} 1) data heterogeneity, where each client may hold data according to a distinct distribution, leading to a sharp drop in accuracy of FL \cite{zhao2018federated}, and 2) heterogeneous clients, which are equipped with a wide range of computation and communication capabilities \textemdash{} challenges the underlying assumption of conventional FL setting that local models have to share the same architecture as the global model \cite{diao2020heterofl}. In the last five years, data heterogeneity has been largely explored in many studies~\cite{karimireddy2019scaffold,lin2020ensemble,li2020federated,seo2020federated,acar2020federated,zhu2021data,li2021model,tan2022towards}. However, only a few works aim to address the problem of heterogeneous clients, particularly memory heterogeneity in FL ~\cite{diao2020heterofl,hong2021efficient}.

One solution to heterogeneous clients is to use the smallest model that all clients can train, but this can severely impact FL performance as larger models tend to perform better ~\cite{frankle2018lottery, neyshabur2018role, bubeck2021universal}. Another approach is to prune channels of the global model for each client based on their memory budgets and average the resulting local models to produce a full-size global model ~\cite{diao2020heterofl, hong2021efficient, horvath2021fjord}.  However, such approaches suffer from the issue of under-expression of small-size models, since the reduction in the width of local models can significantly degrade their performance due to fewer parameters ~\cite{frankle2018lottery}. The negative impact of aggregating small-size models in FL is also verified by our case studies in Section \ref{sec:case_studies}. 

Considering the exceptional performance of the full-size model, we aim to provide an algorithmic solution to enable each client to train the same full-size model and acquire adequate global information in FL. Specifically, we propose memory-adaptive depth-wise learning, where each client sequentially trains blocks of a neural network based on the local memory budget until the full-size model is updated. 
To ensure the classifier layer's supervised signal can be utilized for training each block, we propose two learning strategies: 1) incorporating a skip connection between training blocks and the classifier, and 2) introducing auxiliary classifiers. Our method is suitable for memory-constrained settings as it does not require storing the full intermediate activation and computing full intermediate gradients. Additionally, it can be seamlessly integrated with most FL algorithms, e.g., FedAvg \cite{mcmahan2017communication} and FedProx \cite{li2020federated}.

Apart from providing adaptive strategies for low-memory local training, we investigate the potential of mutual knowledge distillation \cite{hinton2015distilling, zhang2018deep} to address on-the-fly device upgrades or participation of new devices with increased memory capacity. Lastly, we consider devices with extremely limited memory budgets such that some blocks resulting from the finest network decomposition cannot be trained. We propose a partial training strategy, where some blocks that are close to the input sides are never trained throughout.
The main contributions of our work are summarized as follows.

\begin{enumerate}
    \setlength\itemsep{0pt}
    \item Through comprehensive analysis of memory consumption, we develop two memory-efficient training paradigms that empower each client to train a full-size model for improving the global model's performance.
    \item Our framework is model- and optimizer-agnostic. The flexibility allows for deployment in real-world cross-device applications, accommodating clients with varying memory budgets on the fly. 
    \item Our proposed approach is not sensitive to client participation resulting from unstable communication because we learn a unified model instead of different local models as in prior works.
    \item Experimental results 
    demonstrate that the performance of the proposed methods is better than other FL baselines regarding top-1 accuracy in scenarios with heterogeneous memory constraints and diverse non-IID data distributions. We also show the negative impact of sub-networks using width-slimming techniques.
\end{enumerate}

\section{Related Work}
\subsection{Federated Learning}

FL emerges as an important paradigm for learning jointly among clients' decentralized data~\cite{konevcny2016federated,mcmahan2017communication,li2020federated,kairouz2021advances,wang2021field}. One major motivation for FL is to protect users' data privacy, where users' raw data are never disclosed to the server and any other participating users~\cite{abadi2016deep,bonawitz2017practical,sun2019can}.
Partly opened by the {\it federated averaging} (FedAvg)~\cite{mcmahan2017communication}, a line of work tackles FL as a distributed optimization problem where the global objective is defined by a weighted combination of clients' local objectives~\cite{mohri2019agnostic,li2020federated,reddi2020adaptive,wang2020tackling}. The federated learning paradigm of FedAvg has been extended and modified to support different global model aggregation methods and different local optimization objectives and optimizers~\cite{yurochkin2019bayesian,reddi2020adaptive,wang2021field,wang2021local,wang2020federated}. Theoretical analysis has been conducted, which demonstrated that federated optimization enjoys convergence guarantees under certain assumptions~\cite{li2020convergence,wang2021field}.



\subsection{Device Heterogeneity in Federated Learning.}

Especially for cross-device FL, it is a natural setting that client devices are with heterogeneous computation power, communication bandwidth, and/or memory capacity. A few research efforts have been paid to designing memory heterogeneity-aware FL algorithms. HeteroFL~\cite{diao2020heterofl} and FjORD~\cite{horvath2021fjord} allows model architecture heterogeneity among participating clients via varying model widths. The method bears similarity to previously proposed {\it slimmable neural network} ~\cite{yu2018slimmable,yu2019universally} where sub-networks with various widths and shared weights are jointly trained with self-distillation~\cite{zhang2021self}. SplitMix~\cite{hong2021efficient} tackles the same device heterogeneity problem via learning a set of base sub-networks of different sizes among clients based on their hardware capacities, which are later aggregated on-demand according to inference requirements. While recent studies like InclusiveFL \cite{liu2022no} and DepthFL \cite{kim2023depthfl} have also embraced a layer-wise training approach, they allocate model sizes to clients primarily based on a fixed network depth, e.g., taking 2 layers as a computation block.  This method does not accurately represent on-device capabilities during network splitting, as layers at varying depths have distinct computation and memory costs.

In summary, prior research efforts mainly focus on partitioning the global model weights among participating users given their hardware resource constraints, which may be imprecisely evaluated. Consequently, each local user only accesses a portion of the global model.
In this work, however, we seek a holistic approach to handling device heterogeneity in FL without the need for partitioning model weights.  Instead, our approach allows each device to train a full-size model in a sequential manner.

\section{Empirical Study}
\subsection{Preliminaries}
This section briefly reviews prior significant and open-sourced works that aim to address heterogeneous clients in FL, including HeteroFL \cite{diao2020heterofl} and SplitMix \cite{hong2021efficient}. We then conduct an extensive analysis of memory consumption of training a neural network, which has not been explored thoroughly in the FL community.

\subsubsection{Width-scaling FL for heterogeneous clients.}
Existing works such as HeteroFL \cite{diao2020heterofl} and SplitMix \cite{hong2021efficient} address memory heterogeneity by pruning a single global model in terms of channels, creating heterogeneous local models. HeteroFL is the first work in FL that tackles memory heterogeneity via the width-scaling approach but still produces a full-size global model. However, HeteroFL suffers from two major limitations: 1) partial model parameters are under-trained because only partial clients and data are accessible for training the full-size model; 2) small models' information tends to be ignored because of their small-scale parameters. SplitMix was then proposed to address these two issues, which first splits a wide neural network into several base sub-networks for increasing accessible training data, then boosts accuracy by mixing base sub-networks.

\subsubsection{Memory consumption analysis.}
\begin{figure}[ht]
    \centering
    \includegraphics[width=0.35\textwidth]{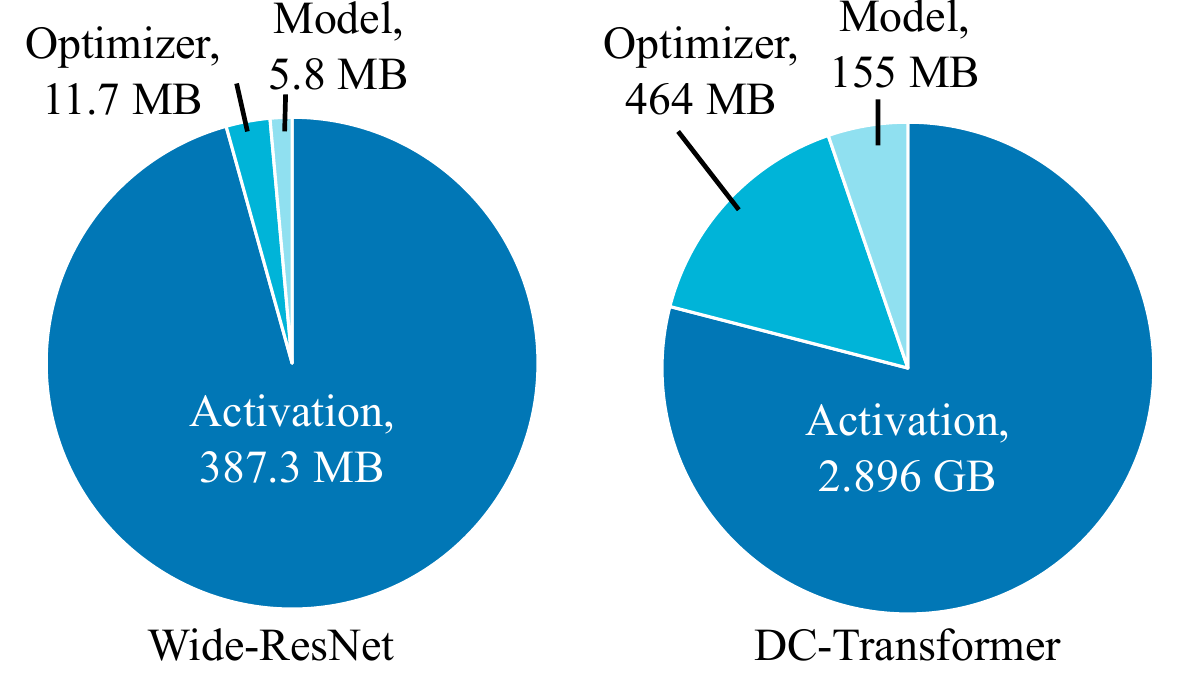}
    \caption{\small{Training memory consumption for {\bf left}: WideResNet on CIFAR-10 and {\bf right}: DC-Transformer on IWSLT’14 German to English. Data source: \cite{sohoni2019low}.}}
    \label{fig:memory_cost}
\end{figure}

~~ Training a neural network with backpropagation consists of feedforward and backward passes \cite{rumelhart1986learning}. 
A feed-forward pass over each block of a neural network generates an activation or output. These intermediate activations are stored in the memory for the backward pass to update the neural network.
Although several works of literature \cite{sohoni2019low, gomez2017reversible, raihan2020sparse, chen2021actnn} demonstrate that activations usually consume most of the memory in standard training of a neural network as shown in Figure \ref{fig:memory_cost}, HeteroFL and SplitMix merely consider the number of model parameters as the memory budget in their experiments. Specifically, they divide clients into groups that are capable of different widths, e.g., a $\frac{1}{8}$-width neural network, which costs approximately $\frac{1}{8}$ activations but only around $\frac{1}{8^2}$ model parameters compared to the full-size neural network. 

\subsection{Behaviors of Sub-networks in Prior Works} \label{sec:case_studies}
In this section, we analyze the behaviors and influences of sub-networks in HeteroFL and SplitMix with respect to the performance of the global model.

\subsubsection{Experimental setting.} We follow the configuration on the CIFAR-10 dataset in SplitMix \cite{hong2021efficient}, where 10 out of 100 clients participate in training in any given communication round, and each client has three classes of the data. In our case studies, we divide clients into four groups with $\{\frac{1}{8}, \frac{1}{4}, \frac{1}{2}, 1\}$-width sub-networks in HeteroFL, and into two groups with $\{r, 1\}$ in SplitMix, where $r = \{\frac{1}{16}, \frac{1}{8}, \frac{1}{4}, \frac{1}{2}\}$. Our observations are summarized below.

\noindent

\begin{figure}[ht]
    \centering
    \includegraphics[width=0.47\textwidth]{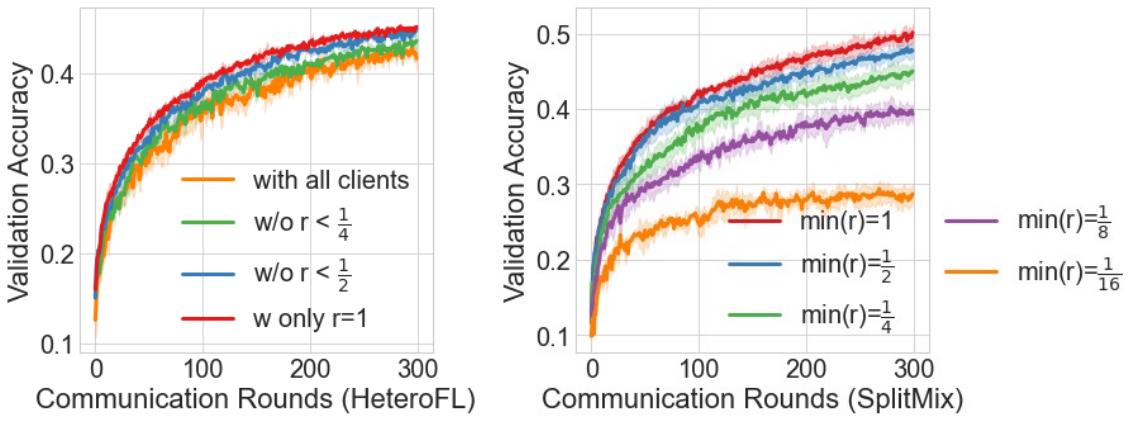}
    \caption{\small{Performance of the global model in HeteroFL {\bf (left)} and SplitMix {\bf (right)} with the varying-width base model, respectively.}}
    \label{fig:motivation}
\end{figure}

\begin{enumerate}
    \setlength\itemsep{0pt}
    \item  \textbf{Small sub-networks make negative contributions in HeteroFL.} Figure \ref{fig:motivation} {\bf (left)} presents typical examples of HeteroFL~\cite{diao2020heterofl} under non-IID settings. The orange line represents the default setting of HeteroFL, where all sub-networks of different widths are aggregated. The other lines indicate specific size of sub-networks that are not aggregated. For example, the green line indicates that the smallest ($\frac{1}{8}$-width) sub-networks do not participate in aggregation. We observe that the global model obtained via aggregating small sub-networks consistently has worse performance than the global model obtained via only aggregating the full-size neural networks, indicating that small size sub-networks make negative contributions.
    
    \item \textbf{Small sub-networks limit global performance in SplitMix.}
    Figure \ref{fig:motivation} {\bf (right)} depicts the prediction performance of the global model in SplitMix \cite{hong2021efficient} by mixing base neural networks with different-width. It clearly illustrated that slimmer base neural networks produce a less powerful global model. Intuitive reasoning is that combining very weak learners leads to an ensemble model with worse generalization.

    \item \textbf{The full-size net makes a difference.}
    Inadequate presence of the full-size models incurs degradation of validation accuracy as shown in Figure \ref{fig:motivation}. Besides, in real-world FL systems, communication can be unstable, and clients with the largest memory budgets may not be available in each round of communication~\cite{bonawitz2017practical}. This constraint limits the practicality of both HeteroFL and SplitMix.
    
\end{enumerate}

\section{Methodology}
Inspired by the observation in the previous section, we introduce a memory-efficient framework \fedepth{} to train full-size neural networks with memory budget constraints in the FL setting. \fedepth{} aims to empirically solve the optimization problem $\min_{\Wcal} F(\Wcal):=\sum_{k=1}^{K} p_k F_k(\Wcal)$. Here, $F_k$ represents the loss function on the $k$th clients. $p_k>0$ for all $k$ and $\sum_{k=1}^{K} p_k=1$. \fedepth{} features memory-adaptive decomposition, where a neural network is decomposed into blocks based on the memory consumption and local clients' memory budgets. An implict assumption made is that all blocks can be trained locally after the decomposition. To further address extreme case that some blocks still cannot fit into the memeory even after the finnest decomposition, \fedepth{} integrates the partial training strategy into the local training stage. We also consider the possibilty that some clients with rich memory budgets may suffer from memory underutilization, hence a variant of \fedepth{} is propsed to use mutual knowledge distillation \cite{zhang2018deep} to boost the performance and fully exploit the local memory of clients.

\subsection{\fedepth{} and Its Variants}\label{sec:fedpthalg}
\begin{algorithm}
\small
\algnewcommand\algorithmicinit{\textbf{Initialization:}}
\algnewcommand\Initialization{\item[\algorithmicinit]}
\algnewcommand\algorithmicoutput{\textbf{Output:}}
\algnewcommand\Output{\item[\algorithmicoutput]}
\caption{\fedepth}
\begin{algorithmic}[1]
    \Require Total number of clients $K$; participation rate $\gamma$; number of communication rounds $R$.
    \Initialization  Model parameter $\Wcal^0$.
    \For{$t=0,\dots,R-1$ communication rounds}
        \State Sample a subset $\Scal^t$ of clients with $|\Scal^t| = \lceil\gamma K\rceil$. 
        \State Broadcast $\Wcal^t$ to clients $k\in\Scal^t$.
        \For {each client $k \in\Scal^t$ \textbf{in parallel}}
            \State $\Wcal^{t+1}_{k} \gets \text{ClientUpdate}(\Wcal^t, k)$.
        \EndFor
        \State Aggregate as $\Wcal^{t+1} = \sum_{k\in\Scal^t}\frac{p_k}{\sum_{k'\in\Scal^t}p_{k'}}\Wcal^{t+1}_{k}.$
    \EndFor
    \Procedure{ClientUpdate}{$\Wcal^{t}, k$}
        \For{$j=1,\cdots,J_k$}
        \State Approximately solve the problem \eqref{eq:jth.subprob}.
        \EndFor
        \State Set $\phi_{k}^{t+1}=\phi_J^{t+1}$
        \State \textbf{Return} $\Wcal^{t}_{k}=\{\theta^{t+1}_{k,1}, \cdots,\theta^{t+1}_{k,J_k}, \phi_{k}^{t+1}\}$.
\EndProcedure
\end{algorithmic}
\label{alg:feddepth}
\end{algorithm}

\subsubsection{Memory-adaptive network decomposition.} 
\begin{figure}[ht]
    \centering
    \includegraphics[scale=0.6]{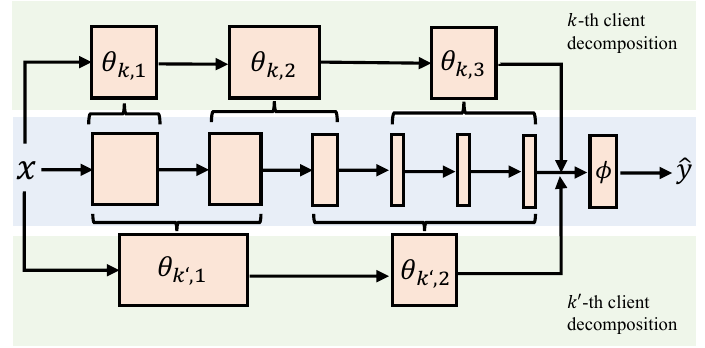}
    \caption{Memory-adaptive neural network decomposition. The second row represents the full neural network with the block size indicating the memory consumption. The first and third rows explain the neural network decomposition based on different clients' memory budgets.}
    \label{fig:decomp.and.flow}
    \vspace{-10pt}
\end{figure}

Since various clients could have drastically different memory budgets, \fedepth{} conducts local training in a memory-adaptive manner. Specifically, for the $k$-th client the full model $\Wcal$ is decomposed to into $J_k+1$ blocks, i.e., $\Wcal=\{\theta_{k,1}, \cdots, \theta_{k,J_k}, \phi\}$, where $\{\theta_{k,j}\}_{j=1}^{J_k}$ and $\phi$ denote body and head of the neural network, respectively. Note that $\theta_{k,j}$ can be different from $\theta_{k',j}$  for any $(k,k',j)$ triple, and the number of parameters contained in $\theta_{k,j}$ is solely determined by the $k$th client's memory budget, hence \fedepth{} is memory-adaptive. In practice, the model decomposition can be determined for each client before training via the estimating memory consumption \cite{gao2020estimating}. See Figure \ref{fig:decomp.and.flow} for an illustration. Suppose the full-size model is composed of 6 layers, where each of layer costs memory of $\{3,2,1,0.5,0.5,0.5\}$ GB, respectively. Assume the $k$th and $k'$th client has 3 GB and 5 GB memory budget, respectively. Then, client $k$ has $J_k = 3$ and client $k'$ has $J_{k'} = 2$ trainable blocks, respectively. That is, client $k'$ will start with training the first two blocks, then the remaining four blocks.

\subsubsection{Depth-wise sequential learning.}
\begin{figure}[ht]
    \centering
    \includegraphics[scale=0.6]{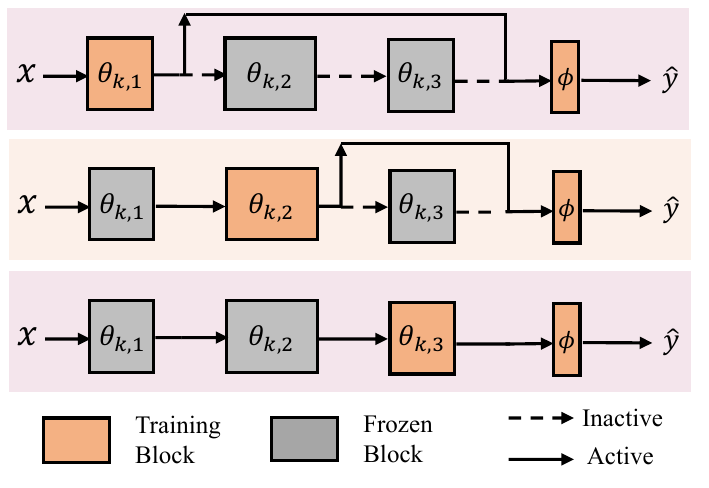}
    \caption{An example of depth-wise sequential learning. There are three training steps: 1) training the first block and the classifier with the skip connection~\cite{he2016deep}; 2) freezing the updated first block and using its activation to train the second block and the classifier with the skip connection; 3) freezing the updated first two blocks and using the activation of the second block to train the third block and the classifier.}
    \label{fig:depth.wise.training}
    \vspace{-10pt}
\end{figure}

Once the decomposition is determined, the $k$-th client at the $t$-th round, locally solves $J_k$ subproblems in a block-wise fashion, i.e., for all $j\in\{1,\dots,J_k\}$,
\begin{equation}\label{eq:jth.subprob}
    (\theta^{t+1}_{k,j}, \phi_j^{t+1}) \in\arg\min_{\theta_{k,j}, \phi}\Lcal(\theta_{k,j}, \phi; \{z_{j-1,i}^{t+1}, y_{i}\}_{i=1}^{n_k}),
\end{equation}
where $\Lcal$ is a loss function, e.g, cross-entropy; $\{z_{j-1,i}^{t+1}\}_{i=1}^{n_k}$ are activations obtained after the local training samples forward-passed through the first $j$ blocks, i.e.,
$z_{j-1,i}^{t+1} = f(x_i;\{\theta_{k,\ell}^{t+1}\}_{\ell=1}^{j-1})$ for $i\in[n_k]$ and $f(\cdot;\{\theta_{k,\ell}^{t+1}\}_{\ell=1}^{j-1})$ is the neural network up to the first $j-1$ blocks; $n_k$ is the total number of training samples. Specifically, $z_{0,i}^{t+1}=x_{i}$ for all $i\in[n_k]$. Problem \eqref{eq:jth.subprob} can be solved by any stochastic gradient-based method. When solving for the $j$-th subproblem at the $t$th round, to fully leverage the global model $\Wcal^t$'s body and the locally newly updated head, we use $(\theta_{j}^{t}, \phi_{j-1}^{t+1})$ as the initial point. See the data flow in Figure \ref{fig:depth.wise.training} when performing the training for the $j$th block. We remark that when the memory budget permits, the activation $\{z_{j-1,i}^{t+1}\}_{i=1}^{n_k}$, which no longer requires the gradient, could be buffered to compute $\{z_{j,i}^{t+1}\}_{i=1}^{n_k}$ once the $\theta_{k,j}^{t+1}$ is obtained, hence saving the redundant forward-pass from the first block to the $j$th block. Also, the number computation required for approximately solving $J_k$ subproblems in the form of \eqref{eq:jth.subprob} should be similar to approximately solving the problem $\min_{\Wcal}F_k(\Wcal)$ if we perform the same number of local updates, and ignore the negligible computation overhead in updating in the head $\phi$. This is because the amount of computation required by one gradient evaluation $\nabla_{\Wcal} F_k$  is equivalent to that of the summation of gradient evaluation of $\{\nabla_{\theta_j} \Lcal\}_{j=1}^{J_k}$.

\subsubsection{Memory-efficient Inference.} 
Depth-wise inference follows the similar logic of the \textit{frozen-then-pass} forward in depth-wise training. Specifically, for each input $x$, we store the activation $z_j$ in the hard drive and discard the predecessor activation $z_{j-1}$. Then we can reload $z_j$ into memory as the input and get the activation $z_{j+1}$. The procedure is repeated until the prediction $\hat y$ is obtained. 

We end this section by giving the detailed algorithmic description in Algorihtm~\ref{alg:feddepth}.

\subsection{Handle Extrem Memory Constraints with Partial Tranining} \label{sec:skip_connection}

\begin{figure}[ht]
    \centering
    \includegraphics[height=4cm]{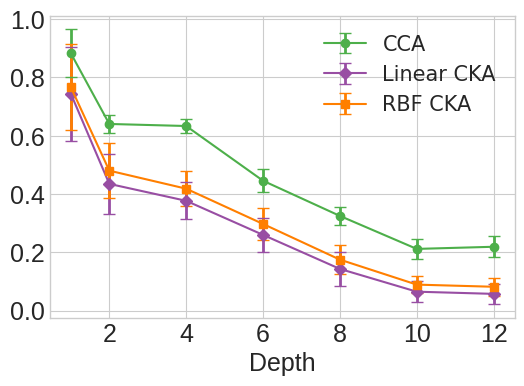}
    \caption{Correspondences between layers of different local neural networks trained from private datasets in FL under non-IID distribution. We observe that early layers, but not later layers, learn similar representations.}
    \label{fig:similarity}
    \vspace{-10pt}
\end{figure}

According to the memory consumption analysis in the Empirical Study, the memory bottleneck of training a neural network is related to the block with the largest activations, which some devices may still not afford. These ``large" blocks are usually the layers close to the input side.
To tackle this issue, we borrow the idea of partial training in \fedepth{}, where we skip the first few blocks that are too large to fit even after the finnest blocks decomposition. This mechanism would not incur significant performance degradation because the input-side layers learn similar representations on different datasets \cite{kornblith2019similarity}, and clients with sufficient memory will provide model parameters of these input-side layers in the FL aggregation phase. Figure \ref{fig:similarity} emeprically validates such a strategy. We train a customized 14-layer ResNet (13 convolution layers and one classifier) on MNIST with 20 clients under non-IID distribution, respectively and measure the similarity of neural network representations, using both canonical correlation analysis (CCA) and centered kernel alignment (CKA) \cite{kornblith2019similarity}.

\subsection{Exploit Sufficient Memory with Mutual Knowledge Distillation}
Previous works on heterogeneous FL ignore the situation where some clients with rich computing resources may participate in the federated training on the fly. Their sufficient memory budget could be potentially utilized to improve the performance of FL via regularizing local model training \cite{mendieta2022local, li2020federated}. Ensembling multiple models is an effective way to improve generalization and reduce variance \cite{shi2021fed, nam2021diversity}. However, considering each model is independently trained in ensemble learning methods, we have to upload/download all of these models in FL settings leading to a significant communication burden. Therefore, we design a new training and aggregation method based on mutual knowledge distillation (MKD)~\cite{hinton2015distilling,zhang2018deep}, where all student neural networks learn collaboratively and teach each other. Therefore, clients with sufficient memory only need to upload one of the local models to the server for aggregation because the knowledge consensus achieved among all models through distillation. 
Formally, assume the $k$th client has a rich memory budget to train $M>1$ models. Then locally it solves 
\small{
\begin{align*}
   \min_{\{\Wcal_{k}^1, \cdots, \Wcal_{k}^M\}}
   & \frac{1}{M}\sum_{m=1}^{M}F_{k}(\Wcal_k^m)  +\frac{1}{M-1} \sum_{m'\neq m}^{M} \KL \left(\mathbf{h}^{m'} \| \mathbf{h}^m\right),
\end{align*}
}
where $\mathbf{h}^m$ are logits calculated from the model $\Wcal_m$ over the local training set and $\KL$ is the Kullback Leibler Divergence. More concretely, $\KL \left(\mathbf{h}^{m'} \| \mathbf{h}^m\right) = \frac{1}{n_k}\sum_{i=1}^{n_k}\KL(h_{i}^{m'} \| h_{i}^{m})$, where $h_{i}^{m}$ is the logits of the $i$th sample computed over model $\Wcal_m$.

\section{Experiments}

\begin{table}[t]
\centering
\small
\begin{tabular}{lccc}
\toprule
\rowcolor{Gray} Depth & Memory & Width & Memory \\ \midrule
B$_{1\sim3}$ & 20.02 &  $\times \frac{1}{8}$ & 14.51 \\
B$_{4}$ & 14.05 & $\times \frac{1}{6}$ & 19.34 \\
B$_{5\sim6}$ & 10.07  & $\times \frac{1}{3}$ & 38.68 \\
B$_{7}$ & 7.21 &  $\times \frac{1}{2}$ & 58.02 \\
 B$_{8\sim9}$ & 5.28 &  $\times 1$ & 116.04 \\ \bottomrule
\end{tabular}
\caption[Caption for LOF]{Memory cost (in MB) with respect to depth and width of PreResNet-20. Each block consists of 2 convolution layers. B$_{1\sim3}$ indicates Block $1,2$ and $3$ in PreResNet-20 have the same memory cost of 20.02 MB. The values are estimated by \textit{pytorch-summary}\protect\footnotemark.} \label{tab:block_memory}
\vspace{-10pt}
\end{table}
\footnotetext{https://github.com/sksq96/pytorch-summary}

\subsection{Experimental Setups}
\noindent \textbf{Datasets and data partition. } 
Our experiments are mainly conducted on CIFAR-10 and CIFAR100 \cite{lecun1998gradient, krizhevsky2009learning}. Extensive results are attached in the appendix. To simulate the non-IID setting with class imbalance, we follow \cite{yurochkin2019bayesian, acar2020federated, gao2022feddc} to distribute each class to clients using the Dirichlet distribution with $\alpha(\lambda)$, where $\lambda = \{0.3, 1.0\}$. Besides, we adopt pathological non-IID data partition $\beta (\Lambda)$ that is used by the selected baselines -- HeteroFL and SplitMix \cite{diao2020heterofl, hong2021efficient}, where each device has unique $\Lambda$ labels with $\Lambda = \{2,5\}$ for CIFAR-10 while $\Lambda = \{10,30\}$ for CIFAR-100. We note that the \textbf{\textit{balanced}} data partition is applied by default, which makes each client holds the same number of examples. The \textbf{\textit{unbalanced}} $\alpha_u (\lambda)$ non-IID, where clients may have a different amount of samples with different feature distribution skew, is also used to evaluate the stability of {\fedepth}.

\noindent \textbf{Memory budgets. } 
Using Pre-Activation ResNet-20 (PreResNet-20)~\cite{he2016identity} as an example, we show the relation of the memory cost of each block between width-wise and depth-wise training in Table \ref{tab:block_memory}. We can see that if clients afford to train $\frac{1}{6}$-width PreResNet-20, they can train the full-size neural network via depth-wise training. The training order is \{ B$_1 \rightarrow$ B$_2 \rightarrow$ B$_3 \rightarrow$ B$_4 \rightarrow$ B$_{5,6}\rightarrow$ B$_{7,8,9}$ \}. Inspired by this example, we simulate three memory budget scenarios. 

\begin{itemize}
    \item \textbf{(Fair).} The memory budgets depend on the hidden channel shrinkage ratio, $r = \{\frac{1}{6}, \frac{1}{3}, \frac{1}{2}, 1\}$, and are uniformly distributed into clients. It means 1/4 of clients can train a PreResNet-20 model within a maximal width of $\frac{1}{6}$-, $\frac{1}{3}$-, $\frac{1}{2}$- and full width, respectively.
    \item \textbf{(Lack).} $r = \{\frac{1}{8}, \frac{1}{6}, \frac{1}{2}, 1\}$. 1/4 of clients with limited memory adopt a partial training strategy.
    \item \textbf{(Surplus).} $r = \{\frac{1}{6}, \frac{1}{3}, \frac{1}{2}, 2\}$. 1/4 of clients with sufficient memory can apply MKD.
\end{itemize}

\begin{table*}[htbp]
\centering
\tiny
\begin{tabular}{ccccccccccc}
\toprule
\multirow{2}{*}{Budget} &
\multirow{2}{*}{Method} & \multicolumn{4}{c}{\cellcolor{Gray} CIFAR-10} & & \multicolumn{4}{c}{\cellcolor{Gray} CIFAR-100} \\  \cmidrule{3-6} \cmidrule{8-11} 
 & & $\alpha(0.3)$ & \multicolumn{1}{c}{$\alpha(1.0)$} & $\beta(2)$ & $\beta(5)$ & & $\alpha(0.3)$ & \multicolumn{1}{c}{$\alpha(1.0)$} & $\beta(10)$ & $\beta(30)$ \\ \midrule
 
 \multirow{1}{*}{\textcolor{gray}{Unrealistic}}
 & \textcolor{gray}{\st{FedAvg ($\times 1$)}} & \textcolor{gray}{\st{63.44 $\pm$ 2.95}} & \textcolor{gray}{\st{69.77 $\pm$ 1.48}} & \textcolor{gray}{\st{34.10 $\pm$ 1.68}} & \textcolor{gray}{\st{68.33 $\pm$ 0.75}} &  & \textcolor{gray}{\st{28.14 $\pm$ 0.48}} & \textcolor{gray}{\st{31.28 $\pm$ 0.17}} & \textcolor{gray}{\st{31.56 $\pm$ 0.28}} & \textcolor{gray}{\st{43.54 $\pm$ 0.16}} \\ \midrule

\multirow{7}{*}{Fair} 
 & FedAvg ($\times \frac{1}{6}$) & 47.66 $\pm$ 2.67 & 53.10 $\pm$ 1.27 & 25.82 $\pm$ 1.34 & 54.02 $\pm$ 0.96 &  & 15.83 $\pm$ 0.25 & 16.39 $\pm$ 0.08 & 13.89 $\pm$ 0.21 & 17.96 $\pm$ 0.08 \\
 & HeteroFL & 56.35 $\pm$ 3.44 & 60.48 $\pm$ 3.23 & 26.47 $\pm$ 2.98 & 57.51 $\pm$ 2.76 &  & 19.20 $\pm$ 3.35 & 23.62 $\pm$ 3.48 & 20.52 $\pm$ 2.14 & 32.04 $\pm$ 2.56  \\
 & SplitMix & 55.63 $\pm$ 2.60 & 58.70 $\pm$ 1.33) & 28.70 $\pm$ 1.92 & 58.15 $\pm$ 2.03 &  & 22.61 $\pm$ 0.88 & 24.86 $\pm$ 0.44 & 19.55 $\pm$ 0.79 & 26.89 $\pm$ 0.32 \\
 & DepthFL & 58.52 $\pm$ 1.76 & 58.55 $\pm$ 1.27 & 29.21 $\pm$ 1.73 & 59.29 $\pm$ 1.60 &  & 23.56 $\pm$ 1.45 & 25.27 $\pm$ 0.94 & 23.14 $\pm$ 1.26 & 33.46 $\pm$ 1.08  \\
 & FeDepth & \colorbox{blue!20}{60.62 $\pm$ 1.84}  & \colorbox{blue!20}{64.95 $\pm$ 1.56} & 30.32 $\pm$ 2.19 & 61.26 $\pm$ 1.83 &  & 26.20 $\pm$ 1.61 & 27.57 $\pm$ 1.67 & 24.47 $\pm$ 1.42 & 36.35 $\pm$ 1.21 \\
 & $m$-FeDepth & 60.03 $\pm$ 1.90 & 65.49 $\pm$ 2.18 & \colorbox{blue!20}{32.50 $\pm$ 2.42} & \colorbox{blue!20}{64.18 $\pm$ 1.95} &  & \colorbox{blue!20}{30.53 $\pm$ 1.30} & \colorbox{blue!20}{32.85 $\pm$ 1.85} & \colorbox{blue!20}{25.15 $\pm$ 1.58} & \colorbox{blue!20}{38.36 $\pm$ 1.34} \\ \cdashlinelr{1-11} 

\multirow{7}{*}{Lack}
& FedAvg ($\times \frac{1}{8}$) & 45.88 $\pm$ 2.92 & 51.30 $\pm$ 1.54 & 23.90 $\pm$ 1.61 & 48.93 $\pm$ 1.23 &  & 14.05 $\pm$ 0.50  & 14.28 $\pm$ 0.32 & 11.62 $\pm$ 0.53 & 15.73 $\pm$ 0.44 \\
& HeteroFL & 54.05 $\pm$ 3.72 & 58.03 $\pm$ 3.58 & 25.83 $\pm$ 3.35 & 55.68 $\pm$ 3.33 &  & 18.40 $\pm$ 3.45  & 21.79 $\pm$ 3.52 & 18.70 $\pm$ 1.34 & 29.33 $\pm$ 2.12 \\
 & SplitMix & 49.18 $\pm$ 2.88 & 52.95 $\pm$ 1.60 & 24.47 $\pm$ 2.29 & 52.40 $\pm$ 1.48 &  & 21.20 $\pm$ 1.23 & 22.60 $\pm$ 1.01 & 17.71 $\pm$ 1.15 & 22.58 $\pm$ 0.64 \\ 
 & DepthFL & 56.04 $\pm$ 2.03 & 57.95 $\pm$ 1.50 & 27.02 $\pm$ 2.12 & 57.13 $\pm$ 2.10 &  & 22.57 $\pm$ 1.68  & 24.18 $\pm$ 1.12 & 21.20 $\pm$ 1.33 & 31.14 $\pm$ 1.45 \\
 & FeDepth & 58.63 $\pm$ 2.11 & \colorbox{red!20}{62.83 $\pm$ 1.83} & \colorbox{red!20}{29.01 $\pm$ 2.54} & 59.76 $\pm$ 1.28 &  & 25.31 $\pm$ 2.52 & 26.97 $\pm$ 1.34 & 22.95 $\pm$ 1.29 & 34.71 $\pm$ 1.78  \\
 & $m$-FeDepth & \colorbox{red!20}{57.56 $\pm$ 2.18} & 62.61 $\pm$ 2.45 & 28.34 $\pm$ 2.79 & \colorbox{red!20}{62.73 $\pm$ 2.44}  &  & \colorbox{red!20}{30.39 $\pm$ 1.68} & \colorbox{red!20}{31.08 $\pm$ 1.57} & \colorbox{red!20}{23.58 $\pm$ 1.41} & \colorbox{red!20}{37.34 $\pm$ 1.89}  \\ \cdashlinelr{1-11} 

\multirow{2}{*}{Surplus} & FeDepth & 61.35 $\pm$ 1.10 & \textbf{67.15 $\pm$ 0.80} & 33.25 $\pm$ 1.80 & 66.17 $\pm$ 1.50 &  & 25.68 $\pm$ 0.85 & 29.85 $\pm$ 1.47 & 27.73 $\pm$ 1.66 & 38.33 $\pm$ 1.53  \\ 
 & $m$-FeDepth & \textbf{62.30 $\pm$ 1.25} & 66.65 $\pm$ 1.65 & \textbf{33.82 $\pm$ 1.85} & \textbf{67.57 $\pm$ 1.89} &  & \textbf{32.55 $\pm$ 1.13} & \textbf{37.00 $\pm$ 1.52} & \textbf{29.20 $\pm$ 1.70} & \textbf{40.42 $\pm$ 1.58} \\ \bottomrule

\end{tabular}
\caption{Test results (top-1 accuracy) under balanced non-IID data partitions using PreResNet-20. Grey texts indicate that the training cannot conform to the pre-defined budget constraint. If not specified, FedAvg denotes the results with \textit{$\times \min(r)$ -width} network. We highlight the best results with \colorbox{blue!20}{Blue Shadow}, \colorbox{red!20}{Red Shadow}, and \textbf{Bold} in the scenarios including clients equipped with fairly sufficient, insufficient and abundant memory, respectively.} \label{tab:balanced} \vspace{-8pt}
\end{table*}

\noindent \textbf{Implementation and evaluation.} 
We compare {\fedepth{}} and its variants with several methods, including FedAvg \cite{mcmahan2017communication}, HeteroFL \cite{diao2020heterofl}, SplitMix \cite{hong2021efficient} and DepthFL \footnote{We reproduced this algorithm to conform to our predefined memory budgets, rather than the original fixed-depth allocation.} \cite{kim2023depthfl} in terms of the average Top-1 Accuracy over 5 different runs. The memory budgets are uniformly distributed to 100 clients. All experiments perform 500 communication rounds with a learning rate of 0.1, local epochs of 10, batch size of 128, SGD optimizer, and a cosine scheduler. 

\subsection{Global Model Evaluation}

\begin{figure}
    \centering
    \includegraphics[scale=0.23]{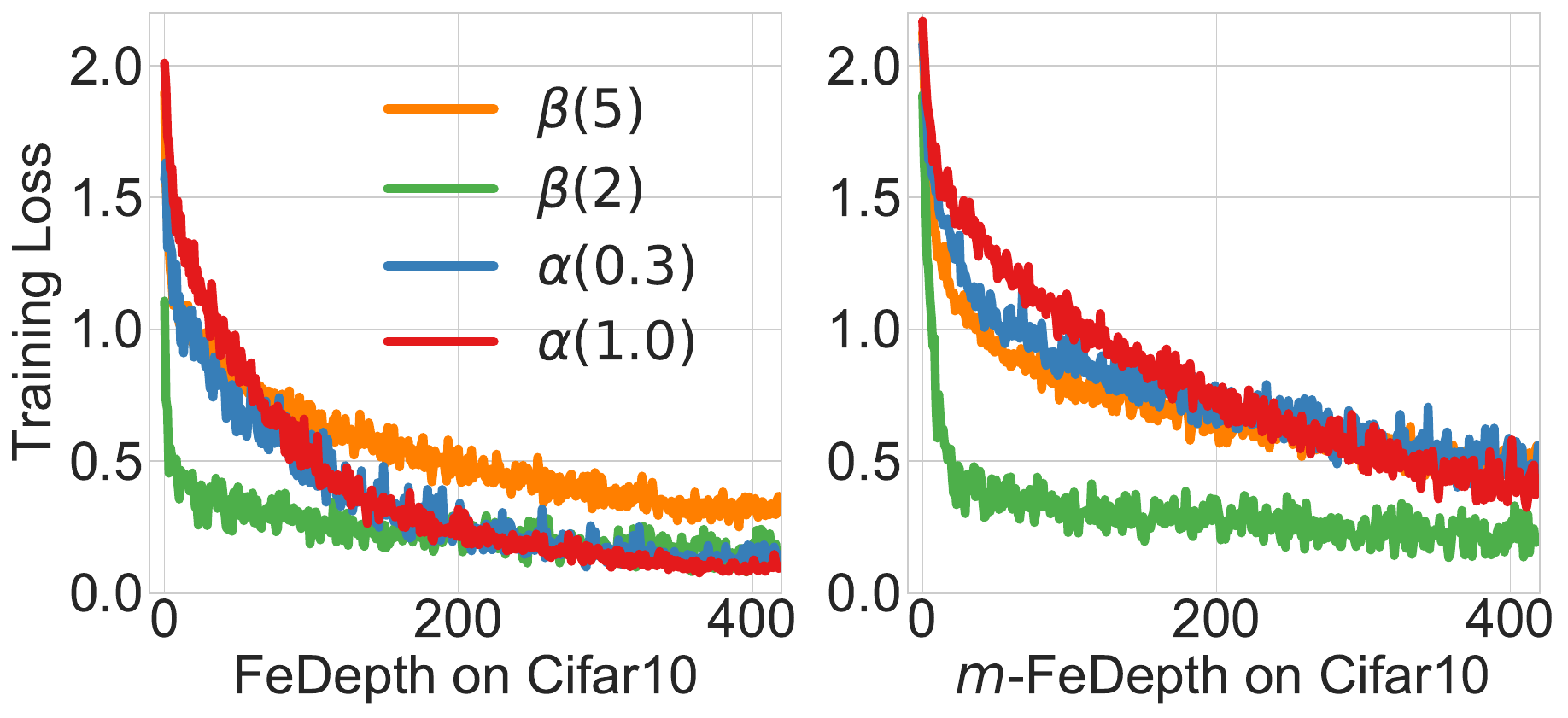}
    \caption{\small{Convergence of \textsc{FeDepth} family on Cifar10.}}
    \label{fig:convergence}
    \vspace{-10pt}
\end{figure}

\noindent \textbf{Results in the \textit{Fair Budget} scenario.} 
In Table \ref{tab:balanced}, we compare test results on a global test dataset (10,000 samples) considering a variety of balanced non-IID data partition and memory constraints. We highlight the best results under different scenarios. In all cases, HeteroFL, SplitMix, and {\fedepth} family outperform vanilla FedAvg, showing their system designs' effectiveness under balanced data distribution. Among all methods, our proposed {\fedepth} and $m$-{\fedepth} achieve the best performance with significant improvements. For example, on CIFAR10, under \textit{Fair Budget}, {\fedepth} gains $~4.09 \pm 0.30 \%$ average improvement compared to HeteroFL while gains $~3.99 \pm 1.77 \%$ average improvement compared to SplitMix. $m$-{\fedepth} gains $~5.35 \pm 1.13 \%$ average improvement compared to HeteroFL while gains $~5.25 \pm 1.20 \%$ average improvement compared to SplitMix. Figure \ref{fig:convergence} shows convergence curves of \fedepth{} on non-IID CIFAR-10 dataset.

\vspace{1mm}
\noindent \textbf{Results in the \textit{Lack Budget} scenario.} 
We observe that HeteroFL has relatively slight accuracy drops or increases compared to the fair budget scenario. The explanation could be deducted from the behaviors of sub-networks discussed in the previous section and Figure \ref{fig:motivation} that small sub-networks slightly influence the global performance because the small number of model parameters provides limited knowledge of the global model in the aggregation phase of FL. In contrast, SplitMix has an apparent performance degradation of an average of $5.55 \pm 0.81 \%$ due to the weaker base learner. The {\fedepth} and $m$-{\fedepth} are relatively stable algorithms against insufficient memory budget, showing $1.73 \pm 0.34 \%$ and $2.74 \pm 0.97 \%$ degradation, respectively.

\vspace{1mm}
\noindent \textbf{Results in the \textit{Surplus Budget} scenario.} We let the new clients with rich resources $r=2$ join in FL and replace the clients with $r=1$. Prior works, like HeteroFL and SplitMix, did not consider such dynamics, and the clients with more memory budgets still train $\times 1$ neural networks. An alternative way is to train a new large base model from scratch and discard previously trained $\times 1$ neural networks, hence wasting computing resources.

From Table \ref{tab:balanced}, we can observe that MKD indeed makes sense for improving the performance of the global model (still $\times 1$-width). Furthermore, we note that combining depth-wise training and MKD is a flexible solution to simultaneously solve dynamic partition, device upgrade, and memory constraints. For example, when a new client with $r = \frac{7}{6}$ enters into the federated learning, the client can locally learn two models via regular and depth-wise sequential training, respectively, and then perform MKD while maintaining an original-size model for aggregation.

\vspace{1mm}
\noindent \textbf{Comparison between \fedepth{} and its variant.} As shown in Table \ref{tab:balanced}, for CIFAR-10, {\fedepth} and $m$-{\fedepth} can achieve similar prediction accuracy. However, for CIFAR-100, $m$-{\fedepth} always outperforms {\fedepth}. It is worth recalling the design of {\fedepth}, which introduces zero paddings to match the dimension between two skip-connected blocks. This may inject negligible noise for the training on more complex data. Replacing the zero paddings with other modules, such as convolutions, may result in a better model. However, this usually comes at the cost of extra memory because of the new activations and parameters, which is usually intolerable to resource-constrained devices. 

\begin{figure*}[ht]
\centering
\begin{minipage}{0.98\textwidth}
    \begin{minipage}[H]{0.48\textwidth}
        \centering
        \includegraphics[width=1\textwidth]{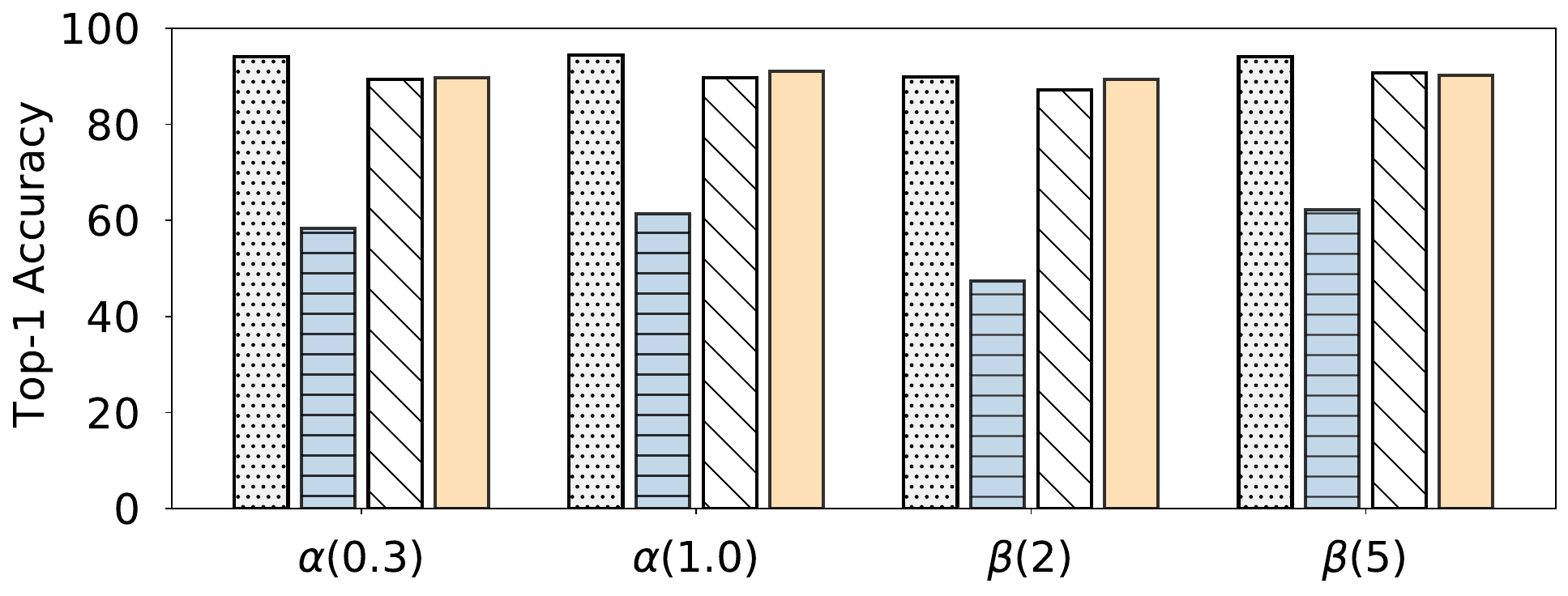}
    \end{minipage}
    \hfill
    \begin{minipage}[H]{0.48\textwidth}
        \centering
        \includegraphics[width=1\textwidth]{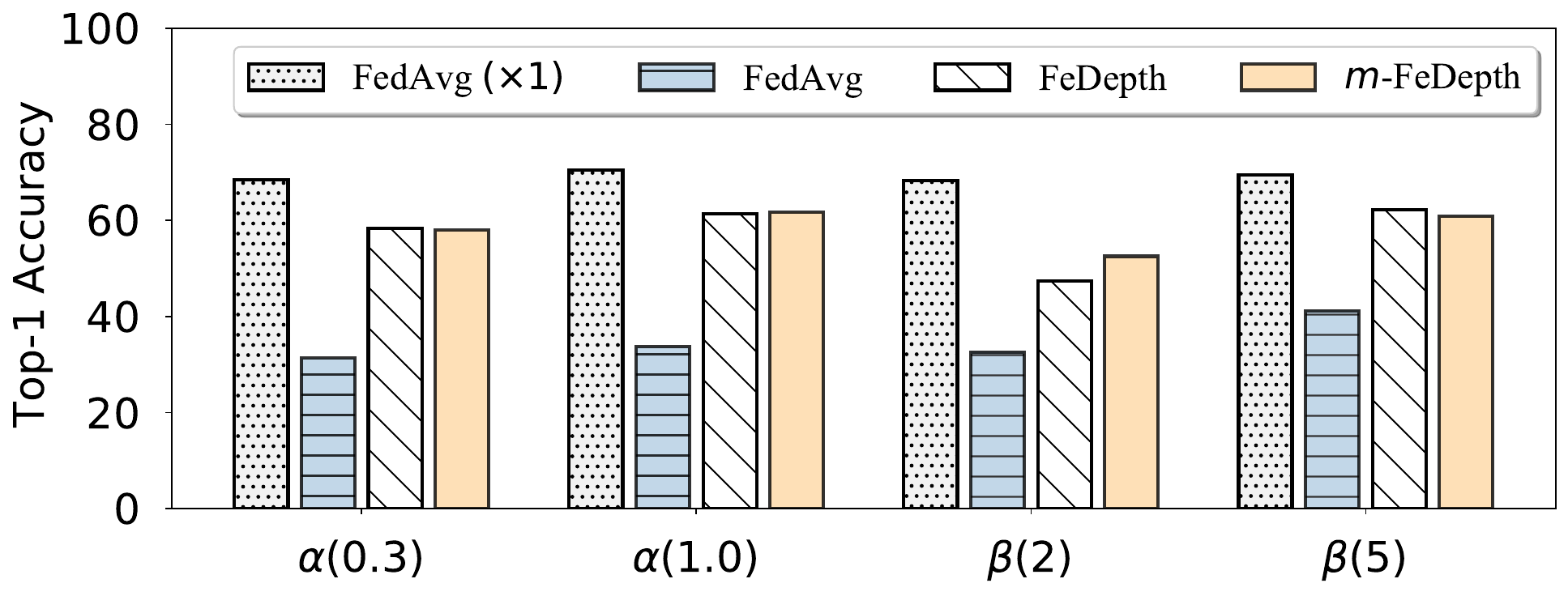}
    \end{minipage}
\end{minipage}
\caption{Fine-tuning ViT-T/16 on CIFAR-10 (\textbf{left}) and CIFAR-100 (\textbf{right}) under balanced non-IID data partitions with FedAvg \fedepth{}, and $m$-\fedepth{}. FedAvg $(\times 1)$ assumes each client can afford to train the full-size model with 12 identical encoder blocks, while FedAvg $(\times \frac{1}{6})$ assumes each client trains a $\frac{1}{6}$-width model, whose memory consumption is equal to train two encoder blocks.} \label{fig:vit}
\vspace{-8pt}
\end{figure*}

\subsection{Influence of Unbalanced Non-IID Distribution} 

Table \ref{tab:unbalanced} shows the prediction results on distributed datasets in FL from the unbalanced Dirichlet partition (\textit{Fair Budget}). We note that HeteroFL and SplitMix were not evaluated on such an unbalanced distribution. Overall, the higher skewed distribution leads to worse performance for FL, which can be observed by comparing results on Table \ref{tab:balanced} and Table \ref{tab:unbalanced}. 

Since the number of samples per class in CIFAR-100 is limited (there are 500 samples for each class), $\alpha_u(\lambda)$ and $\alpha(\lambda)$ will output similar statistical distribution according to the number of samples on each client in FL. Therefore, we obtain similar CIFAR-100 results on both balanced and unbalanced non-IID data partitions. Specifically, $\alpha(\lambda)$ always outputs $400$ training samples per client on average. For CIFAR-100, $\alpha_u(0.3)$ outputs $399.40 \pm 34.53$ training samples per client, $\alpha_u(1.0)$ outputs $399.34 \pm 17.74$. For CIFAR-10, $\alpha_u(0.3)$ outputs $399.44 \pm 150.60$ training samples per client, $\alpha_u(1.0)$ outputs $399.39 \pm 77.37$. 

Regarding CIFAR-10 results, we observe that HeteroFL and SplitMix cannot achieve comparable predictions or generalization ability compared to FedAvg. SplitMix even performs worse than training with the smallest models in FL. This result indicates that SplitMix is not robust to unbalanced distribution. One reason for this phenomenon is that small base models cannot capture representative features due to the significant weight divergence between local clients stemming from a highly skewed distribution \cite{frankle2018lottery, li2022federated}. For HeteroFL, as mentioned in the case study in Section~\ref{sec:case_studies}, the full-size neural networks on resource-sufficient devices provide the fundamental ability but small sub-networks trained with unbalanced distribution indeed affect the global performance. In contrast to HeteroFL and SplitMix, our proposed {\fedepth} and $m$-{\fedepth} gain substantial improvements of $6.15\%$ and $6.53\%$ on CIFAR-10, and of $12.24\%$ and $11.57\%$ on CIFAR-100 compared to FedAvg.

\begin{table}[ht]
\centering
\small
\begin{tabular}{cccccc}
\toprule
   & \multicolumn{2}{c}{\cellcolor{Gray} CIFAR-10} & & \multicolumn{2}{c}{\cellcolor{Gray} CIFAR-100} \\  \cmidrule{2-3} \cmidrule{5-6}  \multirow{-2}{*}{Method}
& $\alpha_u (0.3)$ & $\alpha_u (1.0)$ & & $\alpha_u (0.3)$ & $\alpha_u (1.0)$ \\ \midrule
FedAvg ($\times \frac{1}{6}$) & 46.46 & 52.02 &  & 15.62 & 17.99  \\
HeteroFL & 46.14  & 52.20  &  &  16.02 & 18.36   \\
SplitMix & 31.23 & 44.70 & & 22.68  & 25.28 \\
DepthFL & 47.13 & 55.49 & & 23.19  & 26.02 \\
FeDepth & \textbf{52.61} & \textbf{58.55} &  & 23.25 & 26.16 \\
$m$-FeDepth & 51.58 & 57.91 &  & \textbf{27.86} & \textbf{29.56} \\ \bottomrule
\end{tabular}
\caption{Experimental results on unbalanced Dirichlet partitions. Because of the relatively limited number of samples per class in CIFAR-100, an unbalanced Dirichlet partition outputs similar statistical distribution according to the number of local data examples.} \label{tab:unbalanced}
\vspace{-10 pt}
\end{table}

\subsection{Depth-wise Fine-tuning on ViT}
Foundation models or Transformer architectures \cite{vaswani2017attention, zhou2023comprehensive}, such as Vision Transformer (ViT) \cite{dosovitskiy2020image}, has shown robustness to distribution shifts \cite{bhojanapalli2021understanding}. Recent work has demonstrated that replacing a convolutional network with a pre-trained ViT can greatly accelerate convergence and result in better global models in FL\cite{qu2022rethinking}. Inspired by this finding, we hypothesize that fine-tuning ViT with depth-wise learning will still produce a better global model because 1) decomposing blocks in a depth-wise manner maintains the knowledge learned from pretraining, and 2) memory consumptions of activations in each ViT's block are identical, which indicates that skip connection for handling resource constraints does not introduce any noises and extra parameters. The memory budgets in terms of the width shrinkage ratio $r=\{\frac{1}{6},\frac{1}{3},\frac{1}{2},1\}$ are uniformly allocated to 100 clients as the same setting of PreResNet-20 in the scenario of \textit{Fair Budget}.

For fine-tuning, we choose a learning rate of $5\times 10^4$ and a training epoch of 100. Figure \ref{fig:vit} shows the test results of ViT-T/16 \cite{qu2022rethinking} under balanced Dirichlet data partitions, on which we observe that exploiting $\fedepth$ and $m$-$\fedepth$ can produce good global models. Specifically, $\fedepth$-ViT significantly outperforms $\fedepth$-PreResNet-20 with $36.06 \pm 12.79 \%$ and $30.64 \pm 4.24 \%$ improvements on CIFAR-10 and CIFAR-100 on average, respectively. $m$-$\fedepth$-ViT significantly outperforms $m$-$\fedepth$-PreResNet-20 with $36.66 \pm 12.88 \%$ and $27.41 \pm 3.13 \%$ improvements on CIFAR-10 and CIFAR-100 on average, respectively. We also observe that although local ViTs are fine-tuned on varying distribution data, we obtain global models with similar performance. It indicates that ViT is more robust to distribution shifts and hence improves FL over heterogeneous data. 

\section{Conclusions}
Despite the recent progress in FL, memory heterogeneity still remains largely underexplored. Unlike previous methods based on width-scaling strategies or fixed-depth split, we propose adaptive depth-wise learning for handling varying memory capabilities. The experimental results demonstrate our proposed {\fedepth} family outperform the state-of-the-art algorithms including HeteroFL, SplitMix and DepthFL and are robust to data heterogeneity and client participation. 
Furthermore, using the robustness of ViT to heterogeneous distribution shifts, we reach an excellent global model via our depth-wise solutions.
\fedepth{} is a flexible and scalable framework that can be compatible with most FL algorithms and is reliable to be deployed in practical FL systems and applications.




\bibliography{aaai24}

\begin{thebibliography}{63}
\providecommand{\natexlab}[1]{#1}

\bibitem[{Abadi et~al.(2016)Abadi, Chu, Goodfellow, McMahan, Mironov, Talwar,
  and Zhang}]{abadi2016deep}
Abadi, M.; Chu, A.; Goodfellow, I.; McMahan, H.~B.; Mironov, I.; Talwar, K.;
  and Zhang, L. 2016.
\newblock Deep learning with differential privacy.
\newblock In \emph{Proceedings of the 2016 ACM SIGSAC conference on computer
  and communications security}, 308--318.

\bibitem[{Acar et~al.(2021)Acar, Zhao, Matas, Mattina, Whatmough, and
  Saligrama}]{acar2020federated}
Acar, D. A.~E.; Zhao, Y.; Matas, R.; Mattina, M.; Whatmough, P.; and Saligrama,
  V. 2021.
\newblock Federated Learning Based on Dynamic Regularization.
\newblock In \emph{International Conference on Learning Representations}.

\bibitem[{Bhojanapalli et~al.(2021)Bhojanapalli, Chakrabarti, Glasner, Li,
  Unterthiner, and Veit}]{bhojanapalli2021understanding}
Bhojanapalli, S.; Chakrabarti, A.; Glasner, D.; Li, D.; Unterthiner, T.; and
  Veit, A. 2021.
\newblock Understanding robustness of transformers for image classification.
\newblock In \emph{Proceedings of the IEEE/CVF international conference on
  computer vision}, 10231--10241.

\bibitem[{Bonawitz et~al.(2017)Bonawitz, Ivanov, Kreuter, Marcedone, McMahan,
  Patel, Ramage, Segal, and Seth}]{bonawitz2017practical}
Bonawitz, K.; Ivanov, V.; Kreuter, B.; Marcedone, A.; McMahan, H.~B.; Patel,
  S.; Ramage, D.; Segal, A.; and Seth, K. 2017.
\newblock Practical secure aggregation for privacy-preserving machine learning.
\newblock In \emph{proceedings of the 2017 ACM SIGSAC Conference on Computer
  and Communications Security}, 1175--1191.

\bibitem[{Bubeck and Sellke(2021)}]{bubeck2021universal}
Bubeck, S.; and Sellke, M. 2021.
\newblock A universal law of robustness via isoperimetry.
\newblock \emph{Advances in Neural Information Processing Systems}, 34:
  28811--28822.

\bibitem[{Caldas et~al.(2018)Caldas, Duddu, Wu, Li, Kone{\v{c}}n{\`y}, McMahan,
  Smith, and Talwalkar}]{caldas2018leaf}
Caldas, S.; Duddu, S. M.~K.; Wu, P.; Li, T.; Kone{\v{c}}n{\`y}, J.; McMahan,
  H.~B.; Smith, V.; and Talwalkar, A. 2018.
\newblock Leaf: A benchmark for federated settings.
\newblock \emph{arXiv preprint arXiv:1812.01097}.

\bibitem[{Chen et~al.(2021)Chen, Zheng, Yao, Wang, Stoica, Mahoney, and
  Gonzalez}]{chen2021actnn}
Chen, J.; Zheng, L.; Yao, Z.; Wang, D.; Stoica, I.; Mahoney, M.; and Gonzalez,
  J. 2021.
\newblock Actnn: Reducing training memory footprint via 2-bit activation
  compressed training.
\newblock In \emph{International Conference on Machine Learning}, 1803--1813.
  PMLR.

\bibitem[{Diao, Ding, and Tarokh(2021)}]{diao2020heterofl}
Diao, E.; Ding, J.; and Tarokh, V. 2021.
\newblock HeteroFL: Computation and Communication Efficient Federated Learning
  for Heterogeneous Clients.
\newblock In \emph{International Conference on Learning Representations}.

\bibitem[{Dosovitskiy et~al.(2020)Dosovitskiy, Beyer, Kolesnikov, Weissenborn,
  Zhai, Unterthiner, Dehghani, Minderer, Heigold, Gelly
  et~al.}]{dosovitskiy2020image}
Dosovitskiy, A.; Beyer, L.; Kolesnikov, A.; Weissenborn, D.; Zhai, X.;
  Unterthiner, T.; Dehghani, M.; Minderer, M.; Heigold, G.; Gelly, S.; et~al.
  2020.
\newblock An Image is Worth 16x16 Words: Transformers for Image Recognition at
  Scale.
\newblock In \emph{International Conference on Learning Representations}.

\bibitem[{Du~Terrail et~al.(2022)Du~Terrail, Ayed, Cyffers, Grimberg, He, Loeb,
  Mangold, Marchand, Marfoq, Mushtaq et~al.}]{du2022flamby}
Du~Terrail, J.~O.; Ayed, S.-S.; Cyffers, E.; Grimberg, F.; He, C.; Loeb, R.;
  Mangold, P.; Marchand, T.; Marfoq, O.; Mushtaq, E.; et~al. 2022.
\newblock FLamby: Datasets and Benchmarks for Cross-Silo Federated Learning in
  Realistic Healthcare Settings.
\newblock In \emph{NeurIPS, Datasets and Benchmarks Track}.

\bibitem[{Frankle and Carbin(2019)}]{frankle2018lottery}
Frankle, J.; and Carbin, M. 2019.
\newblock The Lottery Ticket Hypothesis: Finding Sparse, Trainable Neural
  Networks.
\newblock In \emph{International Conference on Learning Representations}.

\bibitem[{Gao et~al.(2022)Gao, Fu, Li, Chen, Xu, and Xu}]{gao2022feddc}
Gao, L.; Fu, H.; Li, L.; Chen, Y.; Xu, M.; and Xu, C.-Z. 2022.
\newblock FedDC: Federated Learning with Non-IID Data via Local Drift
  Decoupling and Correction.
\newblock In \emph{Proceedings of the IEEE/CVF Conference on Computer Vision
  and Pattern Recognition}, 10112--10121.

\bibitem[{Gao et~al.(2020)Gao, Liu, Zhang, Li, Zhu, Lin, and
  Yang}]{gao2020estimating}
Gao, Y.; Liu, Y.; Zhang, H.; Li, Z.; Zhu, Y.; Lin, H.; and Yang, M. 2020.
\newblock Estimating gpu memory consumption of deep learning models.
\newblock In \emph{Proceedings of the 28th ACM Joint Meeting on European
  Software Engineering Conference and Symposium on the Foundations of Software
  Engineering}, 1342--1352.

\bibitem[{Gomez et~al.(2017)Gomez, Ren, Urtasun, and
  Grosse}]{gomez2017reversible}
Gomez, A.~N.; Ren, M.; Urtasun, R.; and Grosse, R.~B. 2017.
\newblock The reversible residual network: Backpropagation without storing
  activations.
\newblock \emph{Advances in neural information processing systems}, 30.

\bibitem[{He et~al.(2016{\natexlab{a}})He, Zhang, Ren, and Sun}]{he2016deep}
He, K.; Zhang, X.; Ren, S.; and Sun, J. 2016{\natexlab{a}}.
\newblock Deep residual learning for image recognition.
\newblock In \emph{Proceedings of the IEEE conference on computer vision and
  pattern recognition}, 770--778.

\bibitem[{He et~al.(2016{\natexlab{b}})He, Zhang, Ren, and
  Sun}]{he2016identity}
He, K.; Zhang, X.; Ren, S.; and Sun, J. 2016{\natexlab{b}}.
\newblock Identity mappings in deep residual networks.
\newblock In \emph{European conference on computer vision}, 630--645. Springer.

\bibitem[{Hinton et~al.(2015)Hinton, Vinyals, Dean
  et~al.}]{hinton2015distilling}
Hinton, G.; Vinyals, O.; Dean, J.; et~al. 2015.
\newblock Distilling the knowledge in a neural network.

\bibitem[{Hong et~al.(2022)Hong, Wang, Wang, and Zhou}]{hong2021efficient}
Hong, J.; Wang, H.; Wang, Z.; and Zhou, J. 2022.
\newblock Efficient Split-Mix Federated Learning for On-Demand and In-Situ
  Customization.
\newblock In \emph{International Conference on Learning Representations}.

\bibitem[{Horvath et~al.(2021)Horvath, Laskaridis, Almeida, Leontiadis,
  Venieris, and Lane}]{horvath2021fjord}
Horvath, S.; Laskaridis, S.; Almeida, M.; Leontiadis, I.; Venieris, S.; and
  Lane, N. 2021.
\newblock Fjord: Fair and accurate federated learning under heterogeneous
  targets with ordered dropout.
\newblock \emph{Advances in Neural Information Processing Systems}, 34:
  12876--12889.

\bibitem[{Kairouz et~al.(2021)Kairouz, McMahan, Avent, Bellet, Bennis, Bhagoji,
  Bonawitz, Charles, Cormode, Cummings et~al.}]{kairouz2021advances}
Kairouz, P.; McMahan, H.~B.; Avent, B.; Bellet, A.; Bennis, M.; Bhagoji, A.~N.;
  Bonawitz, K.; Charles, Z.; Cormode, G.; Cummings, R.; et~al. 2021.
\newblock Advances and open problems in federated learning.
\newblock \emph{Foundations and Trends in Machine Learning}, 14(1-2): 1--210.

\bibitem[{Karimireddy et~al.(2019)Karimireddy, Kale, Mohri, Reddi, Stich, and
  Suresh}]{karimireddy2019scaffold}
Karimireddy, S.~P.; Kale, S.; Mohri, M.; Reddi, S.~J.; Stich, S.~U.; and
  Suresh, A.~T. 2019.
\newblock SCAFFOLD: Stochastic Controlled Averaging for On-Device Federated
  Learning.

\bibitem[{Kim et~al.(2023)Kim, Yu, Kim, and Moon}]{kim2023depthfl}
Kim, M.; Yu, S.; Kim, S.; and Moon, S.-M. 2023.
\newblock Depth{FL} : Depthwise Federated Learning for Heterogeneous Clients.
\newblock In \emph{The Eleventh International Conference on Learning
  Representations}.

\bibitem[{Kone{\v{c}}n{\`y} et~al.(2016)Kone{\v{c}}n{\`y}, McMahan, Yu,
  Richt{\'a}rik, Suresh, and Bacon}]{konevcny2016federated}
Kone{\v{c}}n{\`y}, J.; McMahan, H.~B.; Yu, F.~X.; Richt{\'a}rik, P.; Suresh,
  A.~T.; and Bacon, D. 2016.
\newblock Federated learning: Strategies for improving communication
  efficiency.
\newblock \emph{arXiv preprint arXiv:1610.05492}.

\bibitem[{Kornblith et~al.(2019)Kornblith, Norouzi, Lee, and
  Hinton}]{kornblith2019similarity}
Kornblith, S.; Norouzi, M.; Lee, H.; and Hinton, G. 2019.
\newblock Similarity of neural network representations revisited.
\newblock In \emph{International Conference on Machine Learning}, 3519--3529.
  PMLR.

\bibitem[{Krizhevsky and Hinton(2009)}]{krizhevsky2009learning}
Krizhevsky, A.; and Hinton, G. 2009.
\newblock Learning multiple layers of features from tiny images.

\bibitem[{LeCun et~al.(1998)LeCun, Bottou, Bengio, and
  Haffner}]{lecun1998gradient}
LeCun, Y.; Bottou, L.; Bengio, Y.; and Haffner, P. 1998.
\newblock Gradient-based learning applied to document recognition.
\newblock \emph{Proceedings of the IEEE}, 86(11): 2278--2324.

\bibitem[{Li et~al.(2022)Li, Diao, Chen, and He}]{li2022federated}
Li, Q.; Diao, Y.; Chen, Q.; and He, B. 2022.
\newblock Federated learning on non-iid data silos: An experimental study.
\newblock In \emph{2022 IEEE 38th International Conference on Data Engineering
  (ICDE)}, 965--978. IEEE.

\bibitem[{Li, He, and Song(2021)}]{li2021model}
Li, Q.; He, B.; and Song, D. 2021.
\newblock Model-Contrastive Federated Learning.
\newblock In \emph{Proceedings of the IEEE/CVF Conference on Computer Vision
  and Pattern Recognition}, 10713--10722.

\bibitem[{Li et~al.(2021)Li, Hu, Beirami, and Smith}]{li2021ditto}
Li, T.; Hu, S.; Beirami, A.; and Smith, V. 2021.
\newblock Ditto: Fair and robust federated learning through personalization.
\newblock In \emph{International Conference on Machine Learning}, 6357--6368.
  PMLR.

\bibitem[{Li et~al.(2020{\natexlab{a}})Li, Sahu, Zaheer, Sanjabi, Talwalkar,
  and Smith}]{li2020federated}
Li, T.; Sahu, A.~K.; Zaheer, M.; Sanjabi, M.; Talwalkar, A.; and Smith, V.
  2020{\natexlab{a}}.
\newblock Federated optimization in heterogeneous networks.
\newblock \emph{Proceedings of Machine Learning and Systems}, 2: 429--450.

\bibitem[{Li et~al.(2020{\natexlab{b}})Li, Huang, Yang, Wang, and
  Zhang}]{li2020convergence}
Li, X.; Huang, K.; Yang, W.; Wang, S.; and Zhang, Z. 2020{\natexlab{b}}.
\newblock On the Convergence of FedAvg on Non-IID Data.
\newblock In \emph{International Conference on Learning Representations}.

\bibitem[{Lin et~al.(2020{\natexlab{a}})Lin, Kong, Stich, and
  Jaggi}]{lin2020ensemble}
Lin, T.; Kong, L.; Stich, S.~U.; and Jaggi, M. 2020{\natexlab{a}}.
\newblock Ensemble distillation for robust model fusion in federated learning.
\newblock \emph{Advances in Neural Information Processing Systems}, 33:
  2351--2363.

\bibitem[{Lin et~al.(2020{\natexlab{b}})Lin, Ren, Chen, Ren, Yu, Ma, Rijke, and
  Cheng}]{lin2020meta}
Lin, Y.; Ren, P.; Chen, Z.; Ren, Z.; Yu, D.; Ma, J.; Rijke, M.~d.; and Cheng,
  X. 2020{\natexlab{b}}.
\newblock Meta matrix factorization for federated rating predictions.
\newblock In \emph{Proceedings of the 43rd International ACM SIGIR Conference
  on Research and Development in Information Retrieval}, 981--990.

\bibitem[{Liu et~al.(2022)Liu, Wu, Wu, Wang, Lyu, Chen, and Xie}]{liu2022no}
Liu, R.; Wu, F.; Wu, C.; Wang, Y.; Lyu, L.; Chen, H.; and Xie, X. 2022.
\newblock No one left behind: Inclusive federated learning over heterogeneous
  devices.
\newblock In \emph{Proceedings of the 28th ACM SIGKDD Conference on Knowledge
  Discovery and Data Mining}, 3398--3406.

\bibitem[{McMahan et~al.(2017)McMahan, Moore, Ramage, Hampson, and
  y~Arcas}]{mcmahan2017communication}
McMahan, B.; Moore, E.; Ramage, D.; Hampson, S.; and y~Arcas, B.~A. 2017.
\newblock Communication-efficient learning of deep networks from decentralized
  data.
\newblock In \emph{Artificial intelligence and statistics}, 1273--1282. PMLR.

\bibitem[{Mendieta et~al.(2022)Mendieta, Yang, Wang, Lee, Ding, and
  Chen}]{mendieta2022local}
Mendieta, M.; Yang, T.; Wang, P.; Lee, M.; Ding, Z.; and Chen, C. 2022.
\newblock Local Learning Matters: Rethinking Data Heterogeneity in Federated
  Learning.
\newblock In \emph{Proceedings of the IEEE/CVF Conference on Computer Vision
  and Pattern Recognition}, 8397--8406.

\bibitem[{Mohri, Sivek, and Suresh(2019)}]{mohri2019agnostic}
Mohri, M.; Sivek, G.; and Suresh, A.~T. 2019.
\newblock Agnostic federated learning.
\newblock In \emph{International Conference on Machine Learning}, 4615--4625.
  PMLR.

\bibitem[{Nam et~al.(2021)Nam, Yoon, Lee, and Lee}]{nam2021diversity}
Nam, G.; Yoon, J.; Lee, Y.; and Lee, J. 2021.
\newblock Diversity matters when learning from ensembles.
\newblock \emph{Advances in Neural Information Processing Systems}, 34:
  8367--8377.

\bibitem[{Neyshabur et~al.(2019)Neyshabur, Li, Bhojanapalli, LeCun, and
  Srebro}]{neyshabur2018role}
Neyshabur, B.; Li, Z.; Bhojanapalli, S.; LeCun, Y.; and Srebro, N. 2019.
\newblock The role of over-parametrization in generalization of neural
  networks.
\newblock In \emph{International Conference on Learning Representations}.

\bibitem[{Qu et~al.(2022)Qu, Zhou, Liang, Xia, Wang, Adeli, Fei-Fei, and
  Rubin}]{qu2022rethinking}
Qu, L.; Zhou, Y.; Liang, P.~P.; Xia, Y.; Wang, F.; Adeli, E.; Fei-Fei, L.; and
  Rubin, D. 2022.
\newblock Rethinking architecture design for tackling data heterogeneity in
  federated learning.
\newblock In \emph{Proceedings of the IEEE/CVF Conference on Computer Vision
  and Pattern Recognition}, 10061--10071.

\bibitem[{Raihan and Aamodt(2020)}]{raihan2020sparse}
Raihan, M.~A.; and Aamodt, T. 2020.
\newblock Sparse weight activation training.
\newblock \emph{Advances in Neural Information Processing Systems}, 33:
  15625--15638.

\bibitem[{Reddi et~al.(2020)Reddi, Charles, Zaheer, Garrett, Rush,
  Kone{\v{c}}n{\`y}, Kumar, and McMahan}]{reddi2020adaptive}
Reddi, S.~J.; Charles, Z.; Zaheer, M.; Garrett, Z.; Rush, K.;
  Kone{\v{c}}n{\`y}, J.; Kumar, S.; and McMahan, H.~B. 2020.
\newblock Adaptive Federated Optimization.
\newblock In \emph{International Conference on Learning Representations}.

\bibitem[{Rumelhart, Hinton, and Williams(1986)}]{rumelhart1986learning}
Rumelhart, D.~E.; Hinton, G.~E.; and Williams, R.~J. 1986.
\newblock Learning representations by back-propagating errors.
\newblock \emph{nature}, 323(6088): 533--536.

\bibitem[{Seo et~al.(2020)Seo, Park, Oh, Bennis, and Kim}]{seo2020federated}
Seo, H.; Park, J.; Oh, S.; Bennis, M.; and Kim, S.-L. 2020.
\newblock Federated knowledge distillation.
\newblock \emph{arXiv preprint arXiv:2011.02367}.

\bibitem[{Shi et~al.(2021)Shi, Lai, Kontar, and Chowdhury}]{shi2021fed}
Shi, N.; Lai, F.; Kontar, R.~A.; and Chowdhury, M. 2021.
\newblock Fed-ensemble: Improving generalization through model ensembling in
  federated learning.
\newblock \emph{arXiv preprint arXiv:2107.10663}.

\bibitem[{Sohoni et~al.(2019)Sohoni, Aberger, Leszczynski, Zhang, and
  R{\'e}}]{sohoni2019low}
Sohoni, N.~S.; Aberger, C.~R.; Leszczynski, M.; Zhang, J.; and R{\'e}, C. 2019.
\newblock Low-memory neural network training: A technical report.
\newblock \emph{arXiv preprint arXiv:1904.10631}.

\bibitem[{Sun et~al.(2019)Sun, Kairouz, Suresh, and McMahan}]{sun2019can}
Sun, Z.; Kairouz, P.; Suresh, A.~T.; and McMahan, H.~B. 2019.
\newblock Can you really backdoor federated learning?
\newblock \emph{arXiv preprint arXiv:1911.07963}.

\bibitem[{Tan et~al.(2022)Tan, Yu, Cui, and Yang}]{tan2022towards}
Tan, A.~Z.; Yu, H.; Cui, L.; and Yang, Q. 2022.
\newblock Towards personalized federated learning.
\newblock \emph{IEEE Transactions on Neural Networks and Learning Systems}.

\bibitem[{Vaswani et~al.(2017)Vaswani, Shazeer, Parmar, Uszkoreit, Jones,
  Gomez, Kaiser, and Polosukhin}]{vaswani2017attention}
Vaswani, A.; Shazeer, N.; Parmar, N.; Uszkoreit, J.; Jones, L.; Gomez, A.~N.;
  Kaiser, {\L}.; and Polosukhin, I. 2017.
\newblock Attention is all you need.
\newblock \emph{Advances in neural information processing systems}, 30.

\bibitem[{Wang et~al.(2020{\natexlab{a}})Wang, Yurochkin, Sun, Papailiopoulos,
  and Khazaeni}]{wang2020federated}
Wang, H.; Yurochkin, M.; Sun, Y.; Papailiopoulos, D.; and Khazaeni, Y.
  2020{\natexlab{a}}.
\newblock Federated Learning with Matched Averaging.
\newblock In \emph{International Conference on Learning Representations}.

\bibitem[{Wang et~al.(2021{\natexlab{a}})Wang, Charles, Xu, Joshi, McMahan,
  Al-Shedivat, Andrew, Avestimehr, Daly, Data et~al.}]{wang2021field}
Wang, J.; Charles, Z.; Xu, Z.; Joshi, G.; McMahan, H.~B.; Al-Shedivat, M.;
  Andrew, G.; Avestimehr, S.; Daly, K.; Data, D.; et~al. 2021{\natexlab{a}}.
\newblock A field guide to federated optimization.
\newblock \emph{arXiv preprint arXiv:2107.06917}.

\bibitem[{Wang et~al.(2020{\natexlab{b}})Wang, Liu, Liang, Joshi, and
  Poor}]{wang2020tackling}
Wang, J.; Liu, Q.; Liang, H.; Joshi, G.; and Poor, H.~V. 2020{\natexlab{b}}.
\newblock Tackling the objective inconsistency problem in heterogeneous
  federated optimization.
\newblock \emph{Advances in neural information processing systems}, 33:
  7611--7623.

\bibitem[{Wang et~al.(2021{\natexlab{b}})Wang, Xu, Garrett, Charles, Liu, and
  Joshi}]{wang2021local}
Wang, J.; Xu, Z.; Garrett, Z.; Charles, Z.; Liu, L.; and Joshi, G.
  2021{\natexlab{b}}.
\newblock Local Adaptivity in Federated Learning: Convergence and Consistency.
\newblock \emph{arXiv preprint arXiv:2106.02305}.

\bibitem[{Yu and Huang(2019)}]{yu2019universally}
Yu, J.; and Huang, T.~S. 2019.
\newblock Universally slimmable networks and improved training techniques.
\newblock In \emph{Proceedings of the IEEE/CVF international conference on
  computer vision}, 1803--1811.

\bibitem[{Yu et~al.(2018)Yu, Yang, Xu, Yang, and Huang}]{yu2018slimmable}
Yu, J.; Yang, L.; Xu, N.; Yang, J.; and Huang, T. 2018.
\newblock Slimmable Neural Networks.
\newblock In \emph{International Conference on Learning Representations}.

\bibitem[{Yurochkin et~al.(2019)Yurochkin, Agarwal, Ghosh, Greenewald, Hoang,
  and Khazaeni}]{yurochkin2019bayesian}
Yurochkin, M.; Agarwal, M.; Ghosh, S.; Greenewald, K.; Hoang, N.; and Khazaeni,
  Y. 2019.
\newblock Bayesian nonparametric federated learning of neural networks.
\newblock In \emph{International Conference on Machine Learning}, 7252--7261.
  PMLR.

\bibitem[{Zhang et~al.(2021)Zhang, Jiang, Seversky, Xu, Liu, and
  Song}]{zhang2021federated}
Zhang, K.; Jiang, Y.; Seversky, L.; Xu, C.; Liu, D.; and Song, H. 2021.
\newblock Federated variational learning for anomaly detection in multivariate
  time series.
\newblock In \emph{2021 IEEE International Performance, Computing, and
  Communications Conference (IPCCC)}, 1--9. IEEE.

\bibitem[{Zhang et~al.(2022)Zhang, Wang, Wang, Huang, Yang, Chen, and
  Sun}]{zhang-etal-2022-efficient-federated}
Zhang, K.; Wang, Y.; Wang, H.; Huang, L.; Yang, C.; Chen, X.; and Sun, L. 2022.
\newblock Efficient Federated Learning on Knowledge Graphs via
  Privacy-preserving Relation Embedding Aggregation.
\newblock In \emph{Findings of the Association for Computational Linguistics:
  EMNLP 2022}, 613--621. Abu Dhabi, United Arab Emirates: Association for
  Computational Linguistics.

\bibitem[{Zhang, Bao, and Ma(2021)}]{zhang2021self}
Zhang, L.; Bao, C.; and Ma, K. 2021.
\newblock Self-distillation: Towards efficient and compact neural networks.
\newblock \emph{IEEE Transactions on Pattern Analysis and Machine
  Intelligence}, 44(8): 4388--4403.

\bibitem[{Zhang et~al.(2018)Zhang, Xiang, Hospedales, and Lu}]{zhang2018deep}
Zhang, Y.; Xiang, T.; Hospedales, T.~M.; and Lu, H. 2018.
\newblock Deep mutual learning.
\newblock In \emph{Proceedings of the IEEE conference on computer vision and
  pattern recognition}, 4320--4328.

\bibitem[{Zhao et~al.(2018)Zhao, Li, Lai, Suda, Civin, and
  Chandra}]{zhao2018federated}
Zhao, Y.; Li, M.; Lai, L.; Suda, N.; Civin, D.; and Chandra, V. 2018.
\newblock Federated learning with non-iid data.
\newblock \emph{arXiv preprint arXiv:1806.00582}.

\bibitem[{Zhou et~al.(2023)Zhou, Li, Li, Yu, Liu, Wang, Zhang, Ji, Yan, He
  et~al.}]{zhou2023comprehensive}
Zhou, C.; Li, Q.; Li, C.; Yu, J.; Liu, Y.; Wang, G.; Zhang, K.; Ji, C.; Yan,
  Q.; He, L.; et~al. 2023.
\newblock A Comprehensive Survey on Pretrained Foundation Models: A History
  from BERT to ChatGPT.
\newblock \emph{arXiv preprint arXiv:2302.09419}.

\bibitem[{Zhu, Hong, and Zhou(2021)}]{zhu2021data}
Zhu, Z.; Hong, J.; and Zhou, J. 2021.
\newblock Data-free knowledge distillation for heterogeneous federated
  learning.
\newblock In \emph{International Conference on Machine Learning}, 12878--12889.
  PMLR.

\end{thebibliography}

\clearpage
\begin{figure*}[htbp]
\centering
\begin{minipage}[t]{0.24\textwidth}
\centering
\includegraphics[width=1\textwidth]{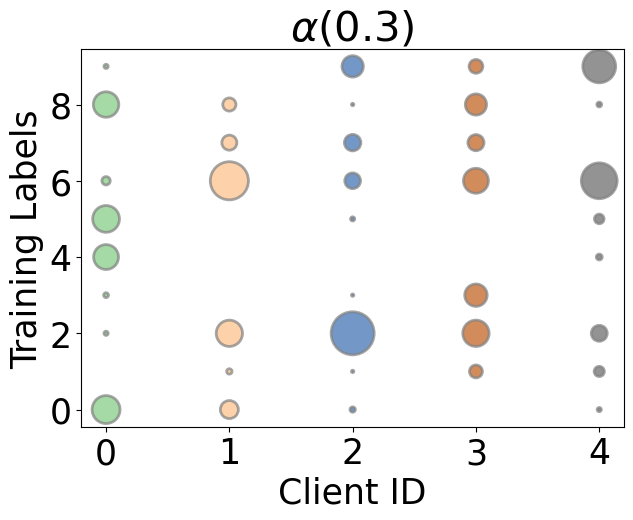}
\end{minipage}
\begin{minipage}[t]{0.24\textwidth}
\centering
\includegraphics[width=1\textwidth]{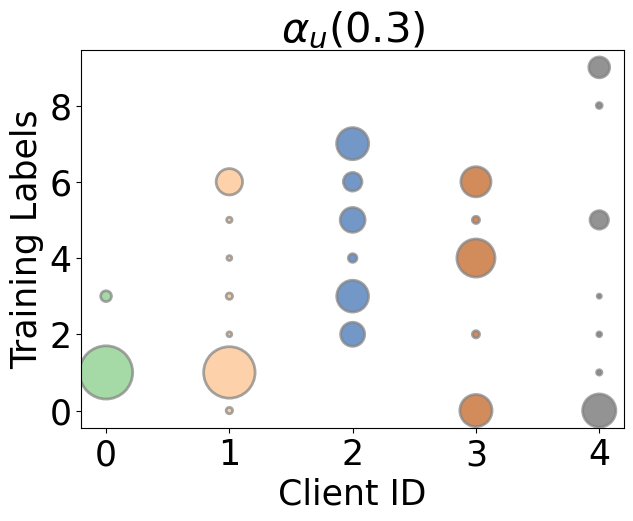}
\end{minipage}
\begin{minipage}[t]{0.24\textwidth}
\centering
\includegraphics[width=1\textwidth]{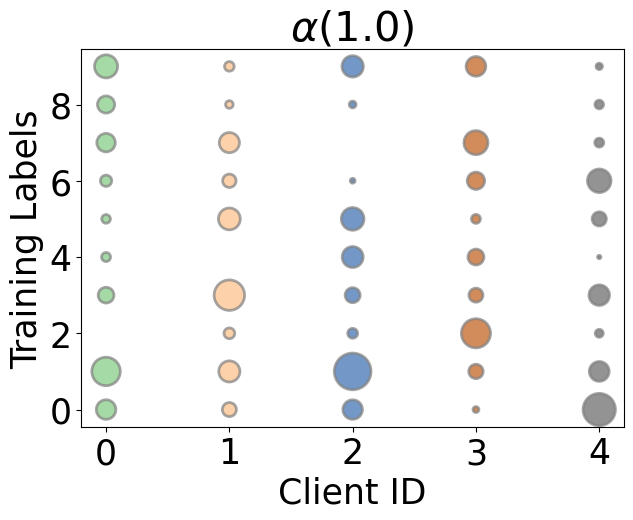}
\end{minipage}
\begin{minipage}[t]{0.24\textwidth}
\centering
\includegraphics[width=1\textwidth]{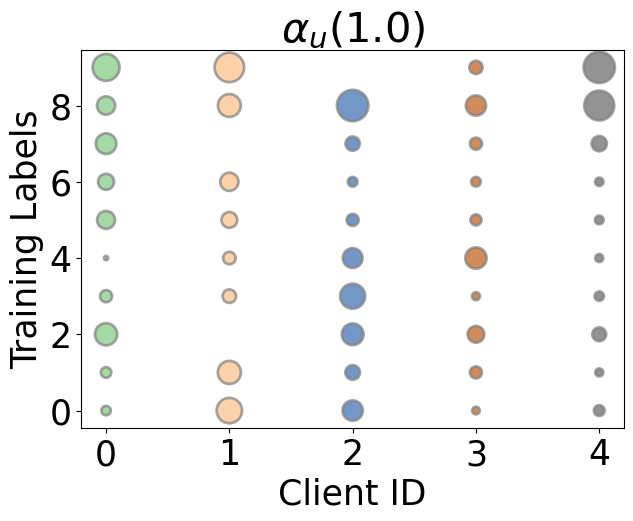}
\end{minipage}
\caption{Visualization of statistical heterogeneity among partial clients on CIFAR-10 dataset, where the $x$-axis indicates client IDs, the $y$-axis indicates class labels, and the size of scattered points indicates the number of training samples for a label available to the client.}
\label{fig:label_distribution}
\vspace{-10pt}
\end{figure*}
\appendix

\section{Visualization of Label Distribution}
We consider 100 clients in all experiments, and in Figure~\ref{fig:label_distribution}, we show label distributions of 5 out of 100 clients under balanced $\alpha(0.3), \alpha(1.0)$ and unbalanced $\alpha_u(0.3), \alpha_u(1.0)$ splits of CIFAR-10.

\section{Extensive Experimental Results}

\subsection{Large-scale FL experiments}
We also conducted experiments on EMNIST with 500 and 1000 clients, respectively, with the 0.1 participation rate, and fair budget with $\alpha(1)$. Additionally, we report the results on FEMNIST \cite{caldas2018leaf}, a \textbf{natural-split FL dataset} derived from partitioning 3597 writers from EMNIST. Furthermore, we present results on TinyImageNet under with 100 clients and 0.1 participation rate. The results are shown in the following table.

\begin{table}[htbp]
  \centering
  {\fontsize{10}{12}\selectfont
    \begin{tabular}{ccccccc}
      \hline
      Datasets & FedAvg & HeteroFL & SplitMix & DepthFL & \textsc{FeDepth} & \textsc{$m$-\textsc{FeDepth}} \\
      \hline
        EMNIST (0.5K) & 77.44 & 79.26 & 70.94 & 73.88 & \textbf{82.07} & 81.73  \\
        EMNIST (1.0K) & 74.24 & 77.95 & 62.04 & 70.18 & \textbf{81.91}  & 81.68 \\
        FEMNIST & 62.69 & 71.05 & 54.62 & 73.80 & \textbf{78.07} & 76.24  \\
        TinyImageNet & 21.00 & 23.98 & 30.87 & 30.89 & 33.97 & \textbf{37.79}\\
      \hline
    \end{tabular}
  }
\end{table}

\subsection{Fairness evaluation}
According to the definition of fairness in FL from \cite{li2021ditto}, we can take the std of test accuracy as a fairness measure. Here we use the std of testing accuracy across 100 clients with the Cifar10 dataset as shown in the following table. Besides, we compare the local training time (in seconds) of each client in one round in the table below.

\begin{table}[htbp]
  \centering
  {\fontsize{10}{12}\selectfont
    \begin{tabular}{ccccccc}
      \hline
      Metric & FedAvg & HeteroFL & SplitMix & \textsc{FeDepth} & \textsc{$m$-\textsc{FeDepth}} \\
      \hline
      Time (s) & 0.42 $\pm$ 0.05 & 0.75 $\pm$ 0.09 & 1.90 $\pm$ 0.60 & 2.49 $\pm$ 0.93 & 2.32 $\pm$ 0.93  \\
      Fairness & 0.05253 & 0.03888 & 0.04919 & 0.04596 & 0.04762 \\
      \hline
    \end{tabular} \label{tab:time}
  }
\end{table}

\end{document}